\documentclass[10pt, a4paper]{article}

\usepackage{lrec-coling2024} 
\usepackage{graphicx}
\usepackage{color, colortbl}
\usepackage{booktabs} 
\usepackage{makecell, multirow}
\usepackage{amsmath,amsfonts}
\usepackage{hyperref}
\usepackage{graphbox}
\usepackage{setspace}
\usepackage{pgfplots}
\usepackage{pgfplotstable}
\usepackage{bm}
\usepackage{xcolor, soul}
\usepackage{tikz}
\definecolor{lightgray}{rgb}{.9,.9,.9}

\newcommand{\hlr}[2][red]{{%
    \colorlet{foo}{#1}%
    \sethlcolor{foo}\hl{#2}}%
}

\title{Efficient and Interpretable Information Retrieval for Product Question Answering with Heterogeneous Data}

\name{Biplob Biswas, Rajiv Ramnath} 

\address{
        The Ohio State University \\
        Columbus OH 43210, USA \\
        \{biswas.102, ramnath.6\}@osu.edu\\
    }

\abstract{
Expansion-enhanced sparse lexical representation improves information retrieval (IR) by minimizing vocabulary mismatch problems during lexical matching. In this paper, we explore the potential of jointly learning dense semantic representation and combining it with the lexical one for ranking candidate information.
We present a hybrid information retrieval mechanism that maximizes lexical and semantic matching while minimizing their shortcomings. Our architecture consists of dual hybrid encoders that independently encode queries and information elements. Each encoder jointly learns a dense semantic representation and a sparse lexical representation augmented by a learnable term expansion of the corresponding text through contrastive learning. We demonstrate the efficacy of our model in single-stage ranking of a benchmark product question-answering dataset containing the typical heterogeneous information available on online product pages. Our evaluation demonstrates that our hybrid approach outperforms independently trained retrievers by 10.95\% (sparse) and 2.7\% (dense) in MRR@5 score. Moreover, our model offers better interpretability and performs comparably to state-of-the-art cross encoders while reducing response time by 30\% (latency) and cutting computational load by approximately 38\% (FLOPs).
 \\ \newline  }

\begin{document}

\maketitleabstract

\section{Introduction}
In the field of natural language processing, ranked information retrieval (IR), refers to retrieving information ordered by relevance from a large collection, in response to a query. Ranked IR remains important even with the emergence of advanced large language models (LLMs) as a means of greatly enriching their outputs. 

Existing retrieval approaches can be categorized into two groups - sparse and dense. Sparse retrieval uses a token-based sparse representation of the query and the information, such as bag-of-words (BoW) obtained via TF-IDF~\cite{tf-idf} or BM25~\cite{bm25}, and an inverted index for query processing. Although these BoW models facilitate faster retrieval, they rely on exact matches, and hence cannot identify semantically relevant information having a different set of tokens than the query. Dense retrieval, on the other hand, retrieves by comparing dense representations often computed by neural networks such as BERT~\cite{devlin2018bert}. While these models can perform semantic-level matching, their computational complexity renders them impractical for online real-time ranking when the corpus becomes large.

In an effort to balance the quality-cost trade-off, a two-stage pipeline is proposed where a quicker retriever first retrieves a smaller set of candidates and then a dense retriever re-ranks them in a second stage. Unfortunately, this approach suffers from two major problems. First, any semantically relevant information pruned due to lack of exact word matches in the first stage is not considered for further ranking. Second, the neural ranker in the last stage lacks interpretability because, for scoring, it uses the inner product of the latent representation of the text which is difficult to explain in human understandable terms.  
Recently proposed transformer~\cite{vaswani2017attention} encoders have the potential to tackle these issues. By utilizing a pre-trained masked language model (MLM), SparTerm~\cite{sparterm} and SPLADE~\cite{splade} progressively improved the use of expansion-aware sparse lexical representation learners in mitigating vocabulary mismatch problems, while enhancing interpretability. SparseEmbed~\cite{sparseembed} further extended this concept by learning contextual embeddings of the top-k tokens in the lexical representation. However, these models ignore the text-level dense representation (i.e. [CLS] token encoding) which captures the summarized expression of a text. Furthermore, being a byproduct of the BERT with MLM head, it can be obtained without additional computation and stored as a single vector. Finally, jointly learning lexical and semantic representations can pave the way for a single-stage ranking, especially in product-question-answering tasks~\cite{hetpqa} where information from an online product page can be pre-computed offline and then ranked at query time.

In this work, we investigate these possibilities and present a hybrid information ranker that balances the quality, cost, and interpretability by incorporating both lexical and semantic matching in ranking. The contribution of our work is in two areas:
\begin{itemize}
    \item We present a hybrid ranking model that jointly learns semantic and lexical representations and combines them for efficient information retrieval.    
    \item We evaluate our model on a heterogeneous product question-answering dataset and show that our approach provides better performance and interpretability with a reasonable computational complexity and memory footprint. Our code is available online\footnote{\url{https://github.com/biplob1ly/HybridPQA}}.
\end{itemize}

\begin{figure*}[!ht]
    \centering
    \includegraphics[width=0.9\textwidth]{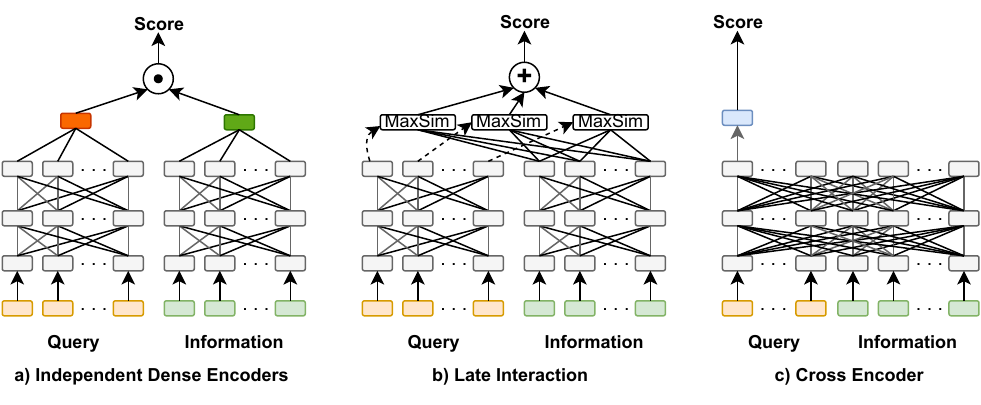}
    \vspace{2mm}
    \caption{Existing neural rankers with different interaction schemes.}
    \label{fig:interaction_models}
\end{figure*}

\section{Related Works}
\label{sec:related_works}
Our hybrid model brings together ideas from both dense retrieval and sparse retrieval. Based on the scoring process, we find three variants of dense retrievers (as shown in \autoref{fig:interaction_models}) related to our work; all of them employ pre-trained language models to learn dense semantic representations. Nogueira et al.~\cite{nogueira2019passage} used BERT as a \textit{cross-encoder}(\autoref{fig:interaction_models}(c)) where concatenated query-information sequence is processed simultaneously through all-to-all interactions and a binary classifier maps the resultant representation to relevance probability. In DPR~\cite{karpukhin2020dense},  Karpukhin et al. employed two \textit{independent dense encoders} (\autoref{fig:interaction_models}(a)) that separately map query and information into their single-vector dense representations and the information score is computed by their inner product. To improve model expressiveness, Khattab et al. proposed ColBERT~\cite{colbert}, a \textit{late-interaction}(\autoref{fig:interaction_models}(b)) model, to utilize a multi-vector representation from dual encoders that allow deferred cross interaction among contextual token encodings. However, ColBERT suffers from scalability issues as it requires storing and indexing all the token encodings in a sequence.

Term-based BM25~\cite{bm25} has been long used as a baseline for sparse retrieval. In order to capture semantic relationships in sparse representations, SNRM~\cite{snrm2018} uses high-dimensional vectors of latent terms. However, it loses the interpretability provided by actual vocabulary terms. SparTerm~\cite{sparterm} addresses this (interpretability) issue by mapping text to a sparse term-importance distribution in BERT vocabulary space. In SPLADE~\cite{splade}, Formal et al. extended this idea by introducing a log-saturation effect in term-importance estimation and sparsity regularization in training loss. Following this, SparseEmbed~\cite{sparseembed} learns and uses contextual embeddings of the sparse lexical representation to improve model expressiveness. Our approach closely follows this direction of research. However, instead of only comparing lexical representation, we also consider summarized semantic matching without increasing encoding complexity, by leveraging the fact that BERT computes the [CLS] token encoding anyway. Moreover, unlike prior hybrid models~\cite{karpukhin2020dense,Ma2021ZeroshotNP,gao2021complement,luan2021sparse}, our model \textit{jointly} learns semantic representations and expandable lexical representations, enabling interpretability with expanded tokens.

\begin{table}
\centering
\scriptsize
\def\arraystretch{1.5}
    \begin{tabular}{l|r|r|r}
    \multicolumn{4}{c}{\textbf{Evidence Ranking}}\\
    \toprule
    \textbf{Items} & \textbf{Train} & \textbf{Validation} & \textbf{Test}\\
    \toprule
    \rowcolor{lightgray}
    \textbf{Total records} & 24295 & 2731 & 309347 \\
    \textbf{Unique query} & 4528 & 509 & 2773 \\
    \rowcolor{lightgray}
    \textbf{Mean candidates per query} &  5.37 &  5.37  &  111.56 \\
    \textbf{Mean +ve candidate ratio} &  0.25  &   0.24  & 0.06 \\
    \rowcolor{lightgray}
    \textbf{Mean question words} &  11.23  &   11.73  & 6.98 \\
    \textbf{Mean candidate words} &  17.19  &   18.49  & 12.59 \\
    \rowcolor{lightgray}
    \textbf{Mean sources per query} &  1.09  &   1.10  & 5.12 \\
    \bottomrule
    \multicolumn{4}{c}{\textbf{Answer Generation}}\\
    \toprule
    \textbf{Items} & \textbf{Train} & \textbf{Validation} & \textbf{Test} \\
    \toprule
    \rowcolor{lightgray}
    \textbf{Total records} & 3693 & 398 & 2289 \\
    \textbf{Unique query} & 3356 & 395 & 1340 \\
    \rowcolor{lightgray}
    \textbf{Mean evidences per query} &  1.1 &  1.01  &  1.71 \\
    \textbf{Mean answer words} &  8.22  &   8.27  & 7.24 \\
    \bottomrule
    \end{tabular}

    \caption{The summary of the hetPQA~\cite{hetpqa} dataset.}
    \label{tab:dataset}
\end{table} 
\section{Dataset}
\label{sec:dataset}
We apply our model to hetPQA~\cite{hetpqa}, a large-scale benchmark dataset for product question-answering systems, that provides various information from product web pages as candidate evidence to answer a product-specific query. In production, after ranking the candidate evidence elements for a query, the higher-ranked ones are utilized for answer generation. The information (evidence) is extracted from heterogeneous sources that include: 1. product attributes in JSON format, 2. bullet points from product summary, 3. community answers to product questions (CQA), 4. product descriptions, 5. on-site publications (OSP) about products, and 6. user reviews on the product page. The collection has separate sets of data for evidence ranking and answer generation, and each dataset comprises train, validation, and test split. The details of the splits are reported in ~\autoref{tab:dataset}. Further, our manual inspection of the BM25-driven evidence ranking result on the test set revealed 1377 incorrect annotations; these were corrected. We have disclosed our correction in the repository shared above and also conducted all our experiments with the amended test set. 
Altogether, the evidence ranking set has 7585 unique questions and 149283 unique pieces of information distributed over the aforementioned 6 sources. The answer generation set contains a total of 5037 unique questions and 5229 unique evidence elements. 
The overall source distribution and average word counts are given in ~\autoref{tab:source_dist}. More details can be found in hetPQA~\cite{hetpqa} paper.

\paragraph{\textbf{Data Preparation}} To begin with, the text was normalized to a canonical representation. All non-English characters were replaced by their equivalents. Symbols and short forms of dimensions (e.g. $3^{\prime\prime}~l \times 4^{\prime\prime}~w$) were substituted by the corresponding English words (\textit{length 3 inches $\times$ width 4 inches}). We also flattened JSON-formatted attributes to comma-separated strings. 

\begin{table}
\centering
\scriptsize
\def\arraystretch{1.5}
\setlength{\tabcolsep}{3pt}
    \begin{tabular}{l|r|r|r|r|r|r}
    \textbf{Items} & \textbf{Attribute} & \textbf{Bullet} & \textbf{CQA} & \textbf{Desc} & \textbf{OSP} & \textbf{Review} \\
    \toprule
    \rowcolor{lightgray}
    \textbf{Ranking} & 4.0\% & 4.4\% & 44.1\% & 12.8\% & 2.6\% & 32.1\% \\
    \textbf{Generation} & 11.4\% & 16.6\% & 21.8\% & 17.8\% & 8.6\% & 23.8\% \\
    \rowcolor{lightgray}
    \textbf{Mean \#words} & 5.8 & 12.6 & 13.3 & 12.9 & 17.8 & 18.4 \\
    \bottomrule
    \end{tabular}
    \vspace{2mm}
    \caption{The distribution of sources and mean word count in the hetPQA~\cite{hetpqa} dataset.}
    \label{tab:source_dist}
\end{table}

\section{Framework}
\label{sec:framework}
Our framework comprises two major components: a ranker and a generator. Given a query and a set of candidate information, the ranker sorts the information in descending order of relevance. The generator then produces a coherent and informative response from the top-ranked results. We elaborate on this in the subsections below.

\begin{figure*}[!ht]
    \centering
    \includegraphics[width=0.9\textwidth]{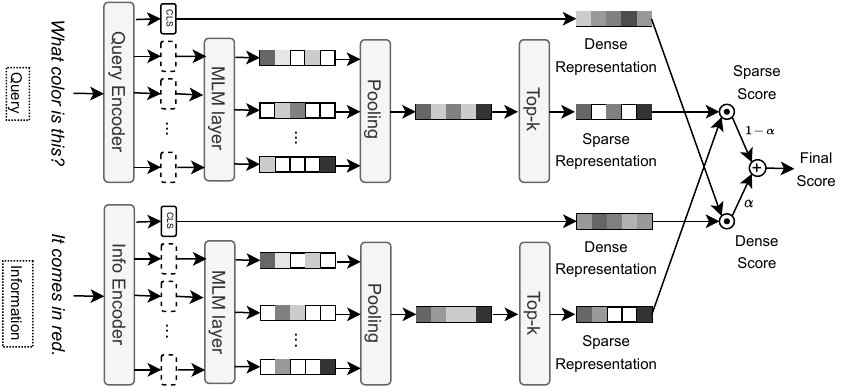}
    \vspace{2mm}
    \caption{The proposed hybrid information ranker.}
    \label{fig:ranker_model}
\end{figure*}

\subsection{Ranker}
The key function of a ranker is to measure the relevance of each candidate information element with respect to the query. 
~\autoref{fig:ranker_model} depicts the architecture of our proposed hybrid ranker. It consists of two separate modules that can independently compute the representations of the queries and information elements.
Given a query $Q=t_{1 \cdots {|Q|}}$ where token $t_i \in V$ for vocabulary $V$, and a candidate information element $C=t_{1 \cdots {|C|}}$ of the same vocabulary, Our ranker first obtains lexical($l$) and semantic($d$) representations of the query and the candidate information as $l_Q$ and $d_Q$, $l_C$ and $d_C$, respectively following the process described in the next subsections. Then the relevance score of the information is computed by the linear interpolation of their semantic and lexical matching:

\begin{equation}
   r(Q, C) = \alpha \times f(d_Q, d_C) + (1 - \alpha) \times f(l_Q, l_C)
\label{eq:rel_score}
\end{equation}
Where $f(Q, C) = Q \cdot C$ and $\alpha \in (0, 1)$ is a hyperparameter indicating importance given to the semantic match.

\subsection{Representation Learning}
\label{sec:repr_learning}
The representation learning procedure for query and information has independent yet similar pipelines as shown in \autoref{fig:ranker_model}. The query-encoder is a pre-trained masked language model (MLM) such as BERT~\cite{devlin2018bert} and it maps the query token sequence to their contextual embeddings $\bm{H_Q} \in \mathbb{R}^{|Q| \times h}$ ($h$: hidden size) and also outputs a summarised representation of entire query in the form of \texttt{[CLS]} token embedding $h_{CLS} \in \mathbb{R}^h$. While the sequence encodings can also be pooled to obtain the summarized vector, it requires additional computation. Instead, we use the pre-trained $h_{CLS}$ as the query's dense semantic representation: $d_Q = h_{CLS}$.

We build on the SPLADE \cite{splade} and SparseEmbed~\cite{sparseembed} methods to compute the lexical representation. In these methods, and as illustrated in \autoref{fig:ranker_model}, the sequence encodings $\bm{H_Q}$ are fed to the BERT's pre-trained MLM head which maps them to MLM logits, $\bm{M_Q} \in \mathbb{R}^{|Q| \times |V|}$. Logit value $m_{i,j}$ in $\bm{M_Q}$ can be considered as an importance indicator of the vocabulary term $v_j \in V$ for the query token $t_i \in Q$. ReLU is applied to the raw logit values to ensure positivity and is followed by a log operation to reduce the dominance of fewer terms. Then the resultant logits are aggregated (using max-pooling or summation) along query token sequences to obtain the combined importance $w_j$ of a term $v_j \in V$ using the following formula:

\begin{equation}
   w_j = \max_{i=1 \dots |Q|}log(1 + ReLU(m_{i,j}))
\label{eq:lex_weights}
\end{equation}

We collect the aggregated importance over the lexical terms, $W=w_{1 \cdots {|V|}}$ through the max pooling layer. To reduce computational complexity during score calculation, we enforce sparsity in $W$ by retaining only the top-k weights in it and zeroing out the rests as shown in ~\autoref{fig:ranker_model}. This leaves us with an expansion-aware sparse lexical representation $l_Q = W$ of the query. 
Following a similar approach for the candidate information element, we obtain its dense semantic representation $d_C$ and its sparse lexical representation $l_C$.

\subsection{Loss Function}
For the training of the hybrid model, we combine ranking loss due to both semantic and lexical representation. Given a dataset $S_{i \cdots |S|}$, where a training instance $S_i$ comprises a query $Q_i$, a piece of positive information $C_i^+$ and $b$ negative candidates $(C_{i,1}^-, C_{i,2}^-, \dots, C_{i,b}^-)$, our model is trained to minimize the following contrastive loss for each kind of representation:
\begin{equation}
    \mathcal{L}_{rank} = -\log \frac{e^{f(Q_i, C_i^+)/\tau}}{e^{f(Q_i, C_i^+)/\tau} + \sum_{j=1}^{b} e^{f(Q_i, C_{i,j}^-)/\tau}}
\end{equation}
Here, $\tau$ is a temperature hyperparameter.
To have an efficient ranking system in terms of computational complexity and memory footprint, it is beneficial to enforce sparsity in the high-dimensional (size: $|V|$) lexical representation. Following SPLADE~\cite{splade}, we also use FLOPS loss for this regularization:
\begin{equation}
    \mathcal{L}_{reg}^C = \sum_{j \in V} \left( \frac{1}{N} \sum_{i=1}^N w_j^{(C_i)}\right)^2
\end{equation}
where $C_i$ is a candidate information element in a batch of size $N$ and $w_j$ is the importance weight of a vocabulary token computed from \autoref{eq:lex_weights}.
Collectively, the training procedure minimizes the following loss function:
\begin{equation}
    \mathcal{L} = \mathcal{L}_{rank}^d + \mathcal{L}_{rank}^l + \lambda^Q\mathcal{L}_{reg}^Q + \lambda^C\mathcal{L}_{reg}^C
\end{equation}
where $\lambda^Q$ and $\lambda^C$ are hyperparameters to introduce higher sparsity in query than information for less scoring cost.

\begin{table*}[ht]
\centering
\resizebox{0.95\textwidth}{!}{
\def\arraystretch{1.5}
\begin{tabular}{c|l|c|c|c|c|c|c}
\toprule
\textbf{\#}
& \textbf{Model}
& \textbf{MAP}
& \textbf{R-Prec}
& \textbf{MRR@5}
& \textbf{NDCG}
& \textbf{Hit Rate@5}
& \textbf{P@1} \\ 
\midrule
a &
\textbf{BM25}~\cite{bm25} &
0.435\hphantom{$^{bcdefghi}$} &
0.388\hphantom{$^{bcdefghi}$} &
0.622\hphantom{$^{bcdefghi}$} &
0.658\hphantom{$^{bcdefghi}$} &
0.796\hphantom{$^{bcdefghi}$} &
0.510\hphantom{$^{bcdefghi}$} \\
b &
\textbf{Cross Encoder}~\cite{nogueira2019passage} &
\textbf{0.604}$^{acdefghi}$\hphantom{} &
\textbf{0.540}$^{acdefghi}$\hphantom{} &
\textbf{0.795}$^{acdefgh}$\hphantom{$^{i}$} &
\textbf{0.780}$^{acdefghi}$\hphantom{} &
\textbf{0.930}$^{acde}$\hphantom{$^{fghi}$} &
\textbf{0.703}$^{acdefgh}$\hphantom{$^{i}$} \\
c &
\textbf{Independent Dense}~\cite{karpukhin2020dense} &
0.552$^{ade}$\hphantom{$^{bfghi}$} &
0.488$^{ae}$\hphantom{$^{bdfghi}$} &
0.761$^{ade}$\hphantom{$^{bfghi}$} &
0.752$^{ade}$\hphantom{$^{bfghi}$} &
0.918$^{ade}$\hphantom{$^{bfghi}$} &
0.659$^{ade}$\hphantom{$^{bfghi}$} \\
d &
\textbf{Late Interaction}\cite{colbert} &
0.544$^{ae}$\hphantom{$^{bcfghi}$} &
0.481$^{ae}$\hphantom{$^{bcfghi}$} &
0.734$^{ae}$\hphantom{$^{bcfghi}$} &
0.741$^{ae}$\hphantom{$^{bcfghi}$} &
0.902$^{ae}$\hphantom{$^{bcfghi}$} &
0.618$^{ae}$\hphantom{$^{bcfghi}$} \\
e &
\textbf{Sparse Lexical}~\cite{splade}, k=128 &
0.505$^{a}$\hphantom{$^{bcdfghi}$} &
0.449$^{a}$\hphantom{$^{bcdfghi}$} &
0.694$^{a}$\hphantom{$^{bcdfghi}$} &
0.713$^{a}$\hphantom{$^{bcdfghi}$} &
0.873$^{a}$\hphantom{$^{bcdfghi}$} &
0.572$^{a}$\hphantom{$^{bcdfghi}$} \\\hline
f &
\textbf{Hybrid, k=128} &
0.563$^{acde}$\hphantom{$^{bghi}$} &
0.498$^{acde}$\hphantom{$^{bghi}$} &
0.770$^{ade}$\hphantom{$^{bcghi}$} &
0.757$^{acde}$\hphantom{$^{bghi}$} &
0.924$^{ade}$\hphantom{$^{bcghi}$} &
0.665$^{ade}$\hphantom{$^{bcghi}$} \\
g &
\textbf{Hybrid, k=256} &
0.572$^{acdef}$\hphantom{$^{bhi}$} &
0.505$^{acdef}$\hphantom{$^{bhi}$} &
0.780$^{acdef}$\hphantom{$^{bhi}$} &
0.763$^{acdef}$\hphantom{$^{bhi}$} &
0.925$^{ade}$\hphantom{$^{bcfhi}$} &
0.679$^{acdef}$\hphantom{$^{bhi}$} \\
h &
\textbf{Hybrid, k=512} &
0.573$^{acdef}$\hphantom{$^{bgi}$} &
0.507$^{acdef}$\hphantom{$^{bgi}$} &
0.782$^{acdef}$\hphantom{$^{bgi}$} &
0.764$^{acdef}$\hphantom{$^{bgi}$} &
0.924$^{ade}$\hphantom{$^{bcfgi}$} &
0.679$^{acdef}$\hphantom{$^{bgi}$} \\
i &
\textbf{Hybrid, k=512, source-aware} &
0.575$^{acdef}$\hphantom{$^{bgh}$} &
0.508$^{acdef}$\hphantom{$^{bgh}$} &
0.792$^{acdefgh}$\hphantom{$^{b}$} &
0.766$^{acdef}$\hphantom{$^{bgh}$} &
0.927$^{acde}$\hphantom{$^{bfgh}$} &
0.697$^{acdefgh}$\hphantom{$^{b}$} \\
\bottomrule
\end{tabular}
}
\vspace{3mm}
\caption{
The overall effectiveness of the experimented rankers on the hetPQA~\cite{hetpqa} dataset.
The best results are highlighted in boldface. Our hybrid model scores are obtained with $\alpha=0.5$.
Superscripts denote significant differences in both Fisher's randomization test and paired Student's t-test with $p \le 0.05$.
}
\label{tab:retrieval_result}
\end{table*}
\subsection{Generator}
Given a query $Q$ and $n$ number of potential information elements $C_{1 \cdots n}$, we aim to generate an answer $A$. To effectively combine multiple information elements for a query, we employ a fusion-in-decoder~\cite{fid2021izacard} model for answer generation. It uses a pre-trained sequence-to-sequence network such as T5~\cite{t52020raffel} that first encodes pairs of question and information $<(Q, C_1), (Q, C_2), \cdots, (Q, C_n)>$ independently and then joins the resultant representations in decoder before performing attention. Finally, we use greedy decoding to generate a natural language answer. As this method processes candidate information elements independently, it allows the aggregation of the elements at a relatively lower latency.

\pgfplotstableread[row sep=\\,col sep=&]{
        source    & BM25 & Cross-Encoder & Independent-Dense & Late-Interaction & Sparse-Lexical & Hybrid \\
        attribute & 0.40 & 0.90 & 0.93 & 0.92 & 0.87 & 0.94\\
        bullet & 0.72 & 0.87 & 0.87 & 0.87 & 0.82 & 0.88\\
        CQA & 0.55 & 0.73 & 0.70 & 0.68 & 0.64 & 0.71\\
        Desc & 0.49 & 0.67 & 0.65 & 0.64 & 0.61 & 0.66\\
        OSP & 0.56 & 0.69 & 0.69 & 0.68 & 0.70 & 0.69\\
        review & 0.65 & 0.79 & 0.75 & 0.73 & 0.71 & 0.77\\
    }\mydata

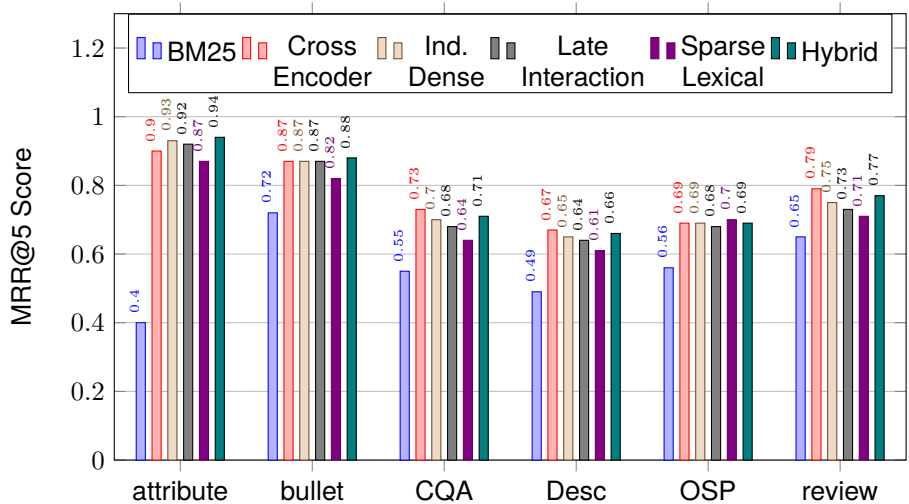
\begin{figure*}
\centering
\begin{tikzpicture}
    \begin{axis}[
            ybar,
            bar width=.12cm,
            ybar=2.5pt,
            width=12cm,
            height=7.5cm,
            legend style={at={(0.5,1)},minimum height=0.9cm,
                anchor=north,legend columns=-1},
            symbolic x coords={attribute, bullet, CQA, Desc, OSP, review, all},
            xtick=data,
            nodes near coords,
            nodes near coords align={vertical},
            nodes near coords style={font=\tiny},
            every node near coord/.append style={rotate=90, anchor=south west,inner ysep=0.5pt},
            ymin=0,ymax=1.3,
            ymajorgrids=true,
            ylabel={MRR@5 Score},
        ]
        \addplot table[x=source,y=BM25]{\mydata};
        \addplot table[x=source,y=Cross-Encoder]{\mydata};
        \addplot table[x=source,y=Independent-Dense]{\mydata};
        \addplot table[x=source,y=Late-Interaction]{\mydata};
        \addplot table[x=source,y=Sparse-Lexical]{\mydata};
        \addplot [fill=teal] table[x=source,y=Hybrid]{\mydata};
        \legend{BM25, \makecell{Cross\\Encoder}, \makecell{Ind.\\Dense}, \makecell{Late\\Interaction}, \makecell{Sparse\\Lexical}, Hybrid}
    \end{axis}
\end{tikzpicture}
\vspace{2mm}
\caption{Ranking results of our hybrid ranker on heterogeneous evidence sources.}
\label{fig:source_ranking}
\end{figure*}
\section{Experimental Environment}
\label{sec:training_details}
We use BERT-base-uncased~\cite{devlin2018bert} (110M parameters) and T5-base~\cite{t52020raffel} (220M parameters) provided by Huggingface~\cite{wolf-etal-2020-transformers} as the core model for evidence ranking and answer generation respectively. We set the following hyperparameters to the relevant models: \{Max token length (each of question, evidence, answer): 128, Warm-up steps: 200, Batch size: 8, Gradient Accumulation Steps: 8, Learning rate: $1e-5$, $\lambda^Q$: $3e-4$, $\lambda^C$: $1e-4$\}. The evidence rankers and generator are trained for 1,500 and 1000 steps respectively and the best checkpoints are considered for evaluation. All the experiments were conducted using a 5-core CPU node at 2.40 GHz, equipped with a single NVIDIA Tesla P100 16GB GPU core and 25 GB of memory. For preprocessing and evaluation, we use NLTK~\cite{bird2009nltk}, calflops~\cite{calflops}, and ranx~\cite{ranx}. Our baseline methods are listed in the first row of \autoref{tab:retrieval_result}. We use Okapi BM25 implementation from rank\_25~\cite{rank_bm25}. For cross-encoder, independent dense encoders, and late-interaction method, we follow the implementation as described in \S\ref{sec:related_works}. The only difference between the sparse-lexical method and our hybrid model is that the former does not incorporate semantic matching in computing the loss and the score. For fairness of comparison, none of our dual-encoders use any additional projection layer on top of BERT's layer and for ranking, we sorted all the candidate information based on \autoref{eq:rel_score} instead of using any indexer.

\section{Evaluation}
In this section, we evaluate the performance of the two modules of our framework, viz., evidence ranking and answer generation.
\label{sec:evaluation}

\subsection{Evidence ranking}
We assess the impact of our proposed method along three dimensions: 1. ranking quality, 2. computational cost and memory footprint, and 3. interpretability. ~\autoref{tab:retrieval_result} lists the experiment results and provides a comparison of our proposed ranking method to the baselines specified in \S\ref{sec:training_details}. Evaluation of all methods was conducted on the amended held-out test set and on the same environment as mentioned in \S\ref{sec:training_details}. There are 2583 unique queries in the test set having at least one positive evidence and we consider only those queries for our evaluation.

\begin{table*}
\centering
\scriptsize
\def\arraystretch{1.5}
    \begin{tabular}{c|c|c|c|c|c|c|c}
    \toprule
    \multicolumn{8}{c}{\textbf{Resource Requirements}}\\
    \bottomrule
    \multicolumn{2}{c|}{\textbf{Metric}} & \textbf{BM25} & \makecell{\textbf{Cross}\\\textbf{Encoder}} & \makecell{\textbf{Independent}\\\textbf{Dense}} & \makecell{\textbf{Late}\\\textbf{Interaction}} & \makecell{\textbf{Sparse}\\\textbf{Lexical}} & \textbf{Hybrid} \\\hline
    \multicolumn{2}{c|}{Params} & - & 109.48M & 2x109.48M & 2x109.48M & 2x109.51M & 2x109.51M \\ \hline
    \multirow{2}{*}{\makecell[c]{Inference\\FLOPs}}
        & Encoding & - & 45.94G & 22.36G & 22.36G & 28.51G & 28.51G \\
        & Interaction & - & - & $2h$ & $2n^2 \cdot h+n$  & $2k$ & $2(h + k)$ \\ \hline
    \multirow{2}{*}{\makecell[c]{Latency\\(ms)}} 
        & Per Query & 0.75 & 475.24 & 229.93 & 275.05 & 296.66 & 331.83 \\
        & Per Info. & 0.007 & 4.2 & 2.04 & 2.43 & 2.63 & 2.93 \\ \hline
    \multicolumn{2}{c|}{\makecell[c]{Offline Storage\\(Per Evidence)}} & - & - & $h$ & $n \cdot h$ & $2k$ & $h+2k$ \\
    \bottomrule
    \end{tabular}
    \vspace{2mm}
    \caption{Resource requirements of the experimented rankers on the hetPQA~\cite{hetpqa} dataset. Here dense representation size $h=768$, max sequence length $n=128$, Count of top tokens considered in lexical representation $k=128$.}
    \label{tab:resource_req}
\end{table*} 
\paragraph{\textbf{Ranking Quality}} To report ranking quality, we utilize six commonly-used evaluation metrics- MAP: mean average precision, R-Prec: precision at the top-R retrieved information elements, MRR@5: mean reciprocal rank within top-5 candidates, NDCG: normalized discounted cumulative gain, Hit rate@5: fraction of queries with at least one positive evidence in top-5 ranked candidates and P@1: precision of the top-ranked evidence. As shown in \autoref{tab:retrieval_result}, the hybrid approach outperformed all other methods except cross-encoder in all metrics. Although the hybrid model with $k=128$ (top token count in lexical representation) bests the independent dense encoder model by a slim margin across the metrics, the difference in their effectiveness becomes statistically significant when more tokens ($k \ge 256$) are considered for lexical matching. The hit-rate@5 indicates the model positions at least one relevant piece of information among the top five in 92.7\% of the queries.

\autoref{fig:source_ranking} illustrates a comparative performance of our proposed method with others across the six different sources of evidence. It shows that our hybrid model dominates existing methods in ranking evidence belonging to the same source. The contrasting score differences between BM25 and neural rankers in attribute and bullet sources not only show the struggle of the pure lexical method with less expressive data but also corroborate the advantage of semantic matching in handling heterogeneous data. In contrast to attribute or bullet evidence which stores clear and concise information, user-driven sources such as CQA and review come with inherent noise including misspellings, presumptive opinions, and so on. According to our manual inspection, these noises contributed to the models' relatively poor performance in these sources.

\pgfplotstableread[row sep=\\, col sep=&, header=true]{
alpha & raw128 & scaled128 & raw256 & scaled256 & raw512 & scaled512\\
0.0 & 0.74 & 0.74 & 0.74 & 0.76 & 0.75 & 0.76\\
0.1 & 0.75 & 0.75 & 0.76 & 0.77 & 0.76 & 0.77\\
0.2 & 0.76 & 0.76 & 0.76 & 0.77 & 0.77 & 0.78\\
0.3 & 0.76 & 0.77 & 0.77 & 0.78 & 0.78 & 0.78\\
0.4 & 0.77 & 0.77 & 0.78 & 0.79 & 0.78 & 0.79\\
0.5 & 0.77 & 0.78 & 0.78 & 0.79 & 0.78 & 0.79\\
0.6 & 0.77 & 0.78 & 0.78 & 0.79 & 0.78 & 0.79\\
0.7 & 0.77 & 0.78 & 0.78 & 0.79 & 0.78 & 0.79\\
0.8 & 0.77 & 0.78 & 0.78 & 0.79 & 0.78 & 0.80\\
0.9 & 0.77 & 0.78 & 0.77 & 0.79 & 0.78 & 0.79\\
1.0 & 0.76 & 0.77 & 0.76 & 0.78 & 0.77 & 0.78\\
}\mydata

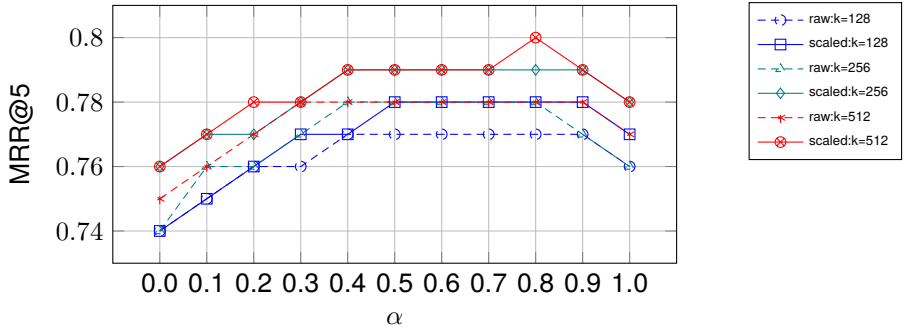
\begin{figure*}[!ht]
\centering
\begin{tikzpicture}
\begin{axis}[
    xtick=data,
    width=9cm,
    height=5cm,
    xticklabels from table={\mydata}{alpha},
    nodes near coords align={vertical},
    ymin=0.73,ymax=0.81,
    ymajorgrids=true,
    xmajorgrids=true,
    legend style={font=\tiny,legend pos=south east,legend cell align=left, at={(1.4,0.4)}},
    xlabel={$\alpha$},
    ylabel={MRR@5},
  ]
\addplot [color=blue, densely dashed, mark=o] table [y=raw128, x expr=\coordindex,] {\mydata};
\addplot [color=blue, mark=square] table [y=scaled128, x expr=\coordindex] {\mydata};
\addplot [color=teal, densely dashed, mark=triangle] table [y=raw256, x expr=\coordindex,] {\mydata};
\addplot [color=teal, mark=diamond] table [y=scaled256, x expr=\coordindex] {\mydata};
\addplot [color=red, densely dashed, mark=asterisk] table [y=raw512, x expr=\coordindex,] {\mydata};
\addplot [color=red, mark=otimes] table [y=scaled512, x expr=\coordindex] {\mydata};
\legend{raw:k=128, scaled:k=128, raw:k=256, scaled:k=256, raw:k=512, scaled:k=512}
\end{axis}
\end{tikzpicture}
\vspace{2mm}
\caption{MRR@5 with regular (dashed) and source-scaled (solid) interaction scores at different semantic and lexical matching combinations.}
\label{fig:score_tuning}
\end{figure*}
\paragraph{\textbf{Resource Requirements}}
\autoref{tab:resource_req} summarizes the resource requirements of our experimented ranking methods. All the methods employ the identical configuration of BERT~\cite{devlin2018bert}. Consequently, the parameter size listed in the first row is roughly proportional to that of a single BERT model except for the sparse-lexical and hybrid models where we have $2 \times 0.03M$ additional parameters for MLM layers. The second row provides the number of floating point operations (FLOPs) needed to be done in the inference stage which includes computation for encoding (measured in Giga-scale: $10^9$) and interaction. Expectedly, the highly performing cross-encoder costs almost double the GFLOPs incurred by the independent dense encoder as the latter only performs query encoding in live and pre-computes the information representation offline. Late-interaction method, on the other hand, is subject to a quadratic interaction cost ($2n^2 \cdot h+ n$) due to its cross term-alignment. In contrast, our hybrid model outperforms all other two-tower rankers (Independent dense, Late-interaction, Sparse lexical) with a moderate $21\%$ increase in encoding GFLOPs and has a linear interaction cost as it only sums the product of the matching query tokens. For latency measurement, we consider the mean combined time elapsed for encoding, interaction, and score-sorting per query as well as per information. Each test set query has an average of 112 information elements. The inference latency is aligned with the inference FLOPs and the latency of our model is halfway between that of the independent dense model and cross encoder.
In terms of offline storage required for each evidence representation, the hybrid approach demands space for a dense vector ($O(h)$) as well as key-value (key: vocabulary token index) pairs corresponding to non-zero elements ($O(k)$) of sparse lexical representation. This memory requirement ($O(h+k)$) is much smaller than that of the late-interaction method ($O(n \cdot h)$) as the latter stores all the token encodings.

\subsection{Ablation Study}
A comparison of evaluation results between our model and models using a subset of components reveals the contribution of additional components in our model. While results in all metrics show a similar trend, we use the standard MRR@5 for our ablation study. To begin with, our hybrid model is of identical architecture as in the sparse lexical model and differs from independent dense models only by the MLM layers. However, our model outperforms the sparse lexical model by 10.95\% and the dense retriever by 0.7\%-2.7\% (for $128 \le k \le 512$) in MRR@5 (\autoref{tab:retrieval_result}). This indicates the benefit of joint learning instead of maximizing only lexical or semantic matching independently.

While our semantic matching captures the underlying summarized meaning, explicit token matching compliments it by allowing us to interpret it. \autoref{fig:score_tuning} illustrates the effect of their contribution on MRR@5 by varying $\alpha$ in \autoref{eq:rel_score} and changing sparsity i.e. the number of top-k tokens (represented by color) considered in sparse representation (see \S\ref{sec:repr_learning}). The differences in area under curves indicate that a higher number of token considerations results in better ranking. On the other hand, the sub-optimal results with lexical-only ($\alpha=0$) or semantic-only ($\alpha=1$) matching and the consistently superior results with $\alpha$ in the range of $0.5-0.8$ further support our hybrid approach. 

\pgfplotstableread[row sep=\\,col sep=&]{
        metric & Copy & Bart-Large & FiD-T5\\
        BLEU & 4 & 34.9 & 39.25 \\
        B-1 & 47.3 & 61.3 & 67 \\
        B-2 & 22.4 & 39.7 & 45 \\
        B-3 & 15.9 & 28.6 & 32.5 \\
        B-4 & 12.6 & 21.6 & 24.2 \\
    }\mydata

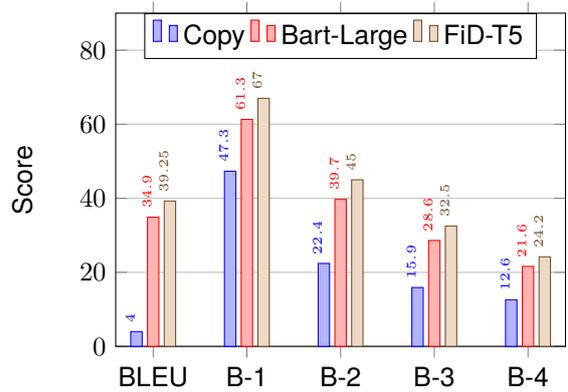
\begin{figure}
\centering
\begin{tikzpicture}
    \begin{axis}[
            ybar,
            bar width=.15cm,
            width=7.5cm,
            height=6cm,
            legend style={at={(0.5,1)},
                anchor=north,legend columns=-1},
            symbolic x coords={BLEU,B-1,B-2,B-3,B-4},
            xtick=data,
            nodes near coords,
            nodes near coords align={vertical},
            nodes near coords style={font=\tiny},
            every node near coord/.append style={rotate=90, anchor=south west,inner ysep=0.5pt},
            ymin=0,ymax=90,
            ymajorgrids=true,
            ylabel={Score},
        ]
        \addplot table[x=metric,y=Copy]{\mydata};
        \addplot table[x=metric,y=Bart-Large]{\mydata};
        \addplot table[x=metric,y=FiD-T5]{\mydata};
        \legend{Copy, Bart-Large, FiD-T5}
    \end{axis}
\end{tikzpicture}
\vspace{2mm}
\caption{Results of answer generation.}
\label{fig:answer_generation}
\end{figure}

Furthermore, our analysis reveals that the distribution of scores across heterogeneous sources differs significantly and favors sources with high mean scores even if they obtain relatively low hit rates at top-5 positions. To counter this, we utilize the source-specific hit-rate@5 obtained from the regular ranking as prior confidence in those sources and multiply it with the combination of normalized scores obtained from \autoref{eq:rel_score}. The resulting ranking score, as shown by the solid lines in \autoref{fig:score_tuning}, outperforms that of the regular ranking in dashed lines by 1\%-3\% across $\alpha$ and $k$ values.

\begin{table*}[t]
\centering
\scriptsize
\renewcommand{\arraystretch}{1.5}
\setlength{\tabcolsep}{0.2em}
    \begin{tabular}{l|l|l|l}
    \toprule
    \textbf{Id} & \textbf{Source} & \textbf{Text} & \textbf{Expansion}\\
    \toprule
    \multirow{3}{*}{\makecell[l]{\#1\\Desc}} 
        & \textbf{Query} & how \hlr[red!41.7]{fast} does the car \hlr[red!44.4]{go}? & \hlr[red!49.2]{speed}, \hlr[red!24.9]{time}\\\cline{2-4}
        & \textbf{Evidence} & maximum \hlr[red!63.9]{speed}: 12 mph & \hlr[red!36.6]{fast}, \hlr[red!33.0]{time}, \hlr[red!30.3]{go}\\\cline{2-4}
        & \textbf{Answer} & \multicolumn{2}{l}{the maximum speed is 12 mph.}\\\hline
    \multirow{3}{*}{\makecell[l]{\#2\\Review}} 
        & \textbf{Query} & how \hlr[red!47.4]{long} do they stay lit? & \hlr[red!37.5]{time}, \hlr[red!30.6]{last}, \hlr[red!24.0]{light}\\\cline{2-4}
        & \textbf{Evidence} & the glow only \hlr[red!56.4]{last}s for on average of 30 minutes. & \hlr[red!54.6]{time}, \hlr[red!56.1]{long}, \hlr[red!48.9]{light}\\\cline{2-4}
        & \textbf{Answer} & \multicolumn{2}{l}{they last under an hour.}\\\hline
    \multirow{3}{*}{\makecell[l]{\#3\\CQA}} 
        & \textbf{Query} & how do you hook it up to a \hlr[red!39.0]{television}? & \hlr[red!45.9]{tv}, \hlr[red!18.3]{power}, \hlr[red!15.6]{plug}\\\cline{2-4}
        & \textbf{Evidence} & you just \hlr[red!61.8]{plug} it directly to your \hlr[red!53.7]{tv}. & \hlr[red!33.3]{power}\\\cline{2-4}
        & \textbf{Answer} & \multicolumn{2}{l}{plug it into your television.}\\\hline
    \multirow{3}{*}{\makecell[l]{\#4\\attribute}} 
        & \textbf{Query} & how \hlr[red!40.8]{tall} is the castle?? & \hlr[red!34.5]{height}, \hlr[red!28.8]{size}\\\cline{2-4}
        & \textbf{Evidence} & item dimensions width: $15.75^{\prime\prime}$, length: $30.5^{\prime\prime}$, \hlr[red!52.2]{height}:$23^{\prime\prime}$ & \hlr[red!36.3]{tall}, \hlr[red!51.3]{size} \\\cline{2-4}
        & \textbf{Answer} & \multicolumn{2}{l}{the castle is 23 inches tall.}\\
    \bottomrule
    \end{tabular}
    \vspace{2mm}
    \caption{Sample evidence prediction and answer generation.}
    \label{tab:samples}
\end{table*} 

\subsection{Generation Quality}
\autoref{fig:answer_generation} illustrates the answer quality generated by three approaches: 1. simply copying the top evidence as an answer, 2. Bart-Large (406M params) and 3. Fusion-in-Decoder with T5 (FiD-T5: 220M params). We utilize the results of the copy-based approach and Bart-Large model from~\cite{hetpqa}. Despite having a smaller number of parameters, the responses generated by FiD-T5 result in a higher BLEU score than that of other approaches. Examples of sample answer generation can be found in \autoref{tab:samples}.

\subsection{Interpretability Analysis - Examples and Discussion} 
A desired quality of a model is to have a simple and human-understandable mechanism to explain its decision-making process. Expanded tokens selected by our model's lexical representations can be interpreted as visualizable faces of underlying thoughts captured in jointly learned semantic representation. Further, the dot product of a matched token importance can be considered as its alignment strength. \autoref{tab:samples} illustrates this idea by highlighting matching tokens of query and predicted evidence. The importance of a token is depicted by its highlighting intensity. The examples demonstrate that our model can match relevant tokens through expansion even if they are not present in the original text. More interestingly, the matching expansion (e.g. \textit{time} in ex\#1, \textit{light} in ex\#2, \textit{power} in ex\#3 and \textit{size} in ex\#4) reveals the shared implicit impression that connects the query and the evidence.

There are a few shortcomings to the model which we leave as future work. First, it treats different forms (e.g. \textit{lasts} and \textit{lasting}) of a root token (e.g. \textit{last}) as separate tokens causing redundant expansion. It can be avoided by merging them with their normalized value. Second, although we reduce the memory footprint of sparse lexical representation by keeping only token index-value pairs, further analysis is required to check its compatibility and efficiency with an indexer such as FAISS~\cite{johnson2019billion}. Without using such an indexer, despite having lower FLOPs, the ranking latency may rise dramatically if we compute token interaction in a loop. Furthermore, differential studies on domain-specific signals such as rate of product sale, count of repeating questions, customers' feedback, and engagement can be measured to quantify the effectiveness of the generator as well as the retriever.

\section{Conclusion}
\label{sec:conclusion}
The study presents a hybrid information ranker that ranks information for a query by comparing their jointly learned dense semantic representations and sparse lexical representations. Our evaluation found that our approach outperformed widely popular sparse or dense retrievers while incurring only a linear cost for both computation and offline storage. Also, our expansion-enhanced lexical matching demonstrates signs of interpretability. In the future, we plan to extend the framework to an end-to-end system with extensive evaluation using a larger dataset.

\nocite{*}
\section{Bibliographical References}\label{sec:reference}

\bibliographystyle{lrec-coling2024-natbib}
\bibliography{hybridPQA}


\end{document}